\documentclass[letterpaper]{article} % DO NOT CHANGE THIS
\usepackage{aaai23}  % DO NOT CHANGE THIS
\usepackage{times}  % DO NOT CHANGE THIS
\usepackage{helvet}  % DO NOT CHANGE THIS
\usepackage{courier}  % DO NOT CHANGE THIS
\usepackage[hyphens]{url}  % DO NOT CHANGE THIS
\usepackage{graphicx} % DO NOT CHANGE THIS
\usepackage{natbib}  % DO NOT CHANGE THIS AND DO NOT ADD ANY OPTIONS TO IT
\usepackage{caption} % DO NOT CHANGE THIS AND DO NOT ADD ANY OPTIONS TO IT
\usepackage{algorithm}
\usepackage{algorithmic}
\usepackage{amsmath}
\usepackage{booktabs}
\usepackage{microtype}
\usepackage{multirow}
\usepackage{xspace}

\usepackage{newfloat}
\usepackage{listings}
\DeclareCaptionStyle{ruled}{labelfont=normalfont,labelsep=colon,strut=off} % DO NOT CHANGE THIS
\floatstyle{ruled}
\newfloat{listing}{tb}{lst}{}
\floatname{listing}{Listing}
\title{Lifelong Embedding Learning and Transfer for Growing Knowledge Graphs}

\author{
    Yuanning Cui\textsuperscript{\rm 1},
    Yuxin Wang\textsuperscript{\rm 1}, 
    Zequn Sun\textsuperscript{\rm 1},
    Wenqiang Liu\textsuperscript{\rm 3},
    Yiqiao Jiang\textsuperscript{\rm 3}, 
    Kexin Han\textsuperscript{\rm 3},
    Wei Hu\textsuperscript{\rm 1,2}\thanks{Corresponding author}
}
\affiliations{
    \textsuperscript{\rm 1} State Key Laboratory for Novel Software Technology, Nanjing University, China\\
    \textsuperscript{\rm 2} National Institute of Healthcare Data Science, Nanjing University, China\\
    \textsuperscript{\rm 3} Interactive Entertainment Group, Tencent Inc, China\\
    \{yncui, zqsun, yxwangcs\}.nju@gmail.com, \{masonqliu, gennyjiang, casseyhan\}@tencent.com, whu@nju.edu.cn
}

\newcommand{\modelname}{LKGE\xspace}

\begin{document}

\maketitle

\begin{abstract}

Existing knowledge graph (KG) embedding models have primarily focused on static KGs. However, real-world KGs do not remain static, but rather evolve and grow in tandem with the development of KG applications. Consequently, new facts and previously unseen entities and relations continually emerge, necessitating an embedding model that can quickly learn and transfer new knowledge through growth. Motivated by this, we delve into an expanding field of KG embedding in this paper, i.e., lifelong KG embedding. We consider knowledge transfer and retention of the learning on growing snapshots of a KG without having to learn embeddings from scratch. The proposed model includes a masked KG autoencoder for embedding learning and update, with an embedding transfer strategy to inject the learned knowledge into the new entity and relation embeddings, and an embedding regularization method to avoid catastrophic forgetting. To investigate the impacts of different aspects of KG growth, we construct four datasets to evaluate the performance of lifelong KG embedding. Experimental results show that the proposed model outperforms the state-of-the-art inductive and lifelong embedding baselines.

\end{abstract}

% ========================>> Introduction
\section{Introduction}
\label{sec:introduction}

\begin{figure}[!t]
\includegraphics[width=\columnwidth]{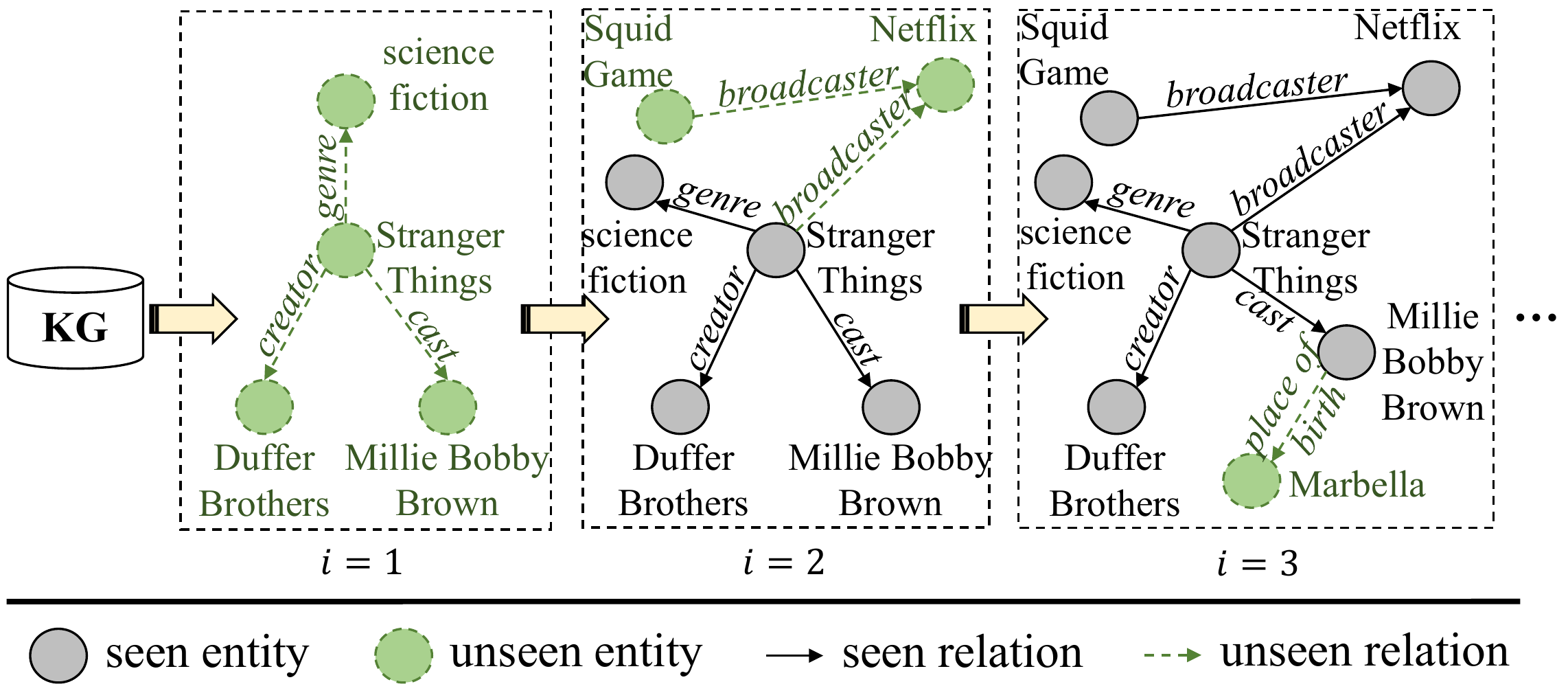}
\caption{An example of a growing KG. In each snapshot $i$, new facts are added into the KG, and previously unseen entities and relations emerge with the new facts.}
\label{fig:illustration}
\vspace{-10pt}
\end{figure}

Many knowledge-driven applications are built on knowledge graphs (KGs), which store massive amounts of structured facts about the real world~\cite{KG_survey}.
Throughout the life-cycle of KG construction and application, 
new facts, unseen entities, and unseen relations continually emerge into the KG on a regular basis. 
As a result, the real-world KG is rarely a static graph, but rather evolves and grows alongside the development.
Figure~\ref{fig:illustration} illustrates an example excerpted from Wikidata~\cite{Wikidata}, which shows the growth of the KG along with the continuous knowledge extraction. 
However, KG embedding, a critical task for downstream applications, has primarily focused on static KGs over the years~\cite{KGE_survey}. 
Learning from scratch every time is inefficient and wastes previously acquired knowledge, and simply fine-tuning new facts would quickly disrupt previously acquired knowledge. 
Hence, this paper proposes to investigate lifelong embedding learning and transfer for growing KGs, with the goal of learning new facts while retaining old knowledge without re-training from scratch.

The key idea of this paper comes from the human learning process.
Humans are typical lifelong learners, with knowledge transfer and retention being the most important aspects of lifelong learning. 
Humans, in particular, can continually learn new knowledge given new facts and use previously learned knowledge to help new knowledge learning (\textit{knowledge learning and transfer}),
as well as update old knowledge while retaining useful knowledge (\textit{knowledge update and retention}). 
Motivated by this, we seek to build a lifelong KG embedding model, namely \textbf{\modelname}, which is capable of learning, transferring and retaining knowledge for growing KGs efficiently.
Existing related work, such as inductive KG embedding~\cite{MEAN,LAN}, mainly focuses on knowledge transfer, ignoring new knowledge learning and old knowledge update.

The proposed lifelong KG embedding task faces two major challenges.
First, how to strike a balance between new knowledge learning and old knowledge transfer? 
Learning embeddings for new entities and relations from scratch cannot leverage previously learned knowledge, and inductively generating embeddings for them ignores new knowledge in the new snapshot.
Second, how to update old knowledge while retaining useful knowledge?
Learning new facts about an old entity usually requires updating the previously learned embeddings, which can be harmful to the old model.
This is because updating an old entity embedding would affect many other old embeddings of related entities.
This would cause the catastrophic forgetting issue, and therefore affect the applications built on the old KG snapshot.

To resolve the above challenges, we propose three solutions in our \modelname model.
First, as the base embedding model for new knowledge learning and old knowledge update, we design a masked KG autoencoder that masks and reconstructs the entities or relations in new facts. 
It builds connections between locally-related old and new entities, acting as a bridge for knowledge transfer.
Second, to aid in the learning of new knowledge, we propose a knowledge embedding transfer strategy that uses previously learned knowledge to initialize the embeddings of new entities and relations.
These embeddings are used by the KG autoencoder to learn new facts.
Third, to avoid catastrophic forgetting in the old knowledge update, 
we propose embedding regularization to balance the learning of new facts and the update of old embeddings.

We build four datasets to assess the lifelong KG embedding performance, including the
entity-centric, relation-centric, fact-centric, and hybrid growth.
Each dataset examines a different aspect of KG growth.
By contrast, existing datasets \cite{MEAN,DiCGRL,CKGE} all assume that a KG grows in an ideal way, with balanced new entities or facts in each new snapshot. 
In our experiments, we compare the link prediction accuracy, knowledge transfer ability, and learning efficiency of the proposed model against baselines on the four datasets. 
The results show that the proposed \modelname not only achieves the best performance on the four datasets, but also has the best forward knowledge transfer ability and learning efficiency.
The main contributions of this paper are summarized as follows:
\begin{itemize}
\item We study \textit{lifelong KG embedding learning and transfer}. It is a practical task since the real-world KGs continually evolve and grow, which requires the embedding model to be capable of handling the knowledge growth.

\item We propose \textit{a novel lifelong learning model}, \modelname. It includes a masked KG autoencoder as the basis of embedding learning and update, an embedding transfer strategy for knowledge transfer, and an embedding regularization to prevent catastrophic forgetting in knowledge update.

\item We conduct \textit{extensive experiments on four new datasets}. The results demonstrate the effectiveness and efficiency of \modelname against a variety of state-of-the-art models.
\end{itemize}

% ========================>> Related Work
\section{Related Work}
\label{sec:related_work}

%In this section, we review two lines of related work, i.e., KG embedding and lifelong learning.

\subsection{Knowledge Graph Embedding}
\label{subsec:knowledge_graph_embedding}

KG embedding seeks to encode the symbolic representations of KGs into vector space to foster the application of KGs in downstream tasks. 
Most existing KG embedding models \cite{TransE,TransH,ConvE,R-GCN,RSN,InteractE,CompGCN} focus on static graphs and cannot continually learn new knowledge on the growing KGs.

To embed unseen entities, inductive KG embedding models learn to represent an entity by aggregating its existing neighbors in the previous KG snapshot. 
MEAN \cite{MEAN} uses a graph convolutional network (GCN) \cite{GCN} for neighborhood aggregation. 
When an unseen entity emerges, the GCN would aggregate its previously seen neighboring entities to generate an embedding. 
LAN \cite{LAN} adopts an attention mechanism to attentively aggregate different neighbors. 
As MEAN and LAN rely on the entity neighborhood for embedding learning, they cannot handle the new entities that have no neighbors in the previous snapshot.
Furthermore, inductive KG embedding disregards learning the facts about new entities.

Our work is also relevant to dynamic KG embedding. 
puTransE \cite{puTransE} trains several new models when facts are added. 
DKGE \cite{DKGE} learns contextual embeddings for entities and relations, which can be automatically updated as the KG grows.
They both need partial re-training on old facts, but our model does not.
In addition, some subgraph-based models, such as GraIL \cite{GraIL}, INDIGO \cite{INDIGO}, and TACT \cite{TACT}, can also represent unseen entities using the entity-independent features and subgraph aggregation.
Their subgraph-building process is time-consuming, making them only applicable to small KGs.
In order to run on large-scale KGs, NBFNet \cite{NBFNet} proposes a fast node pair embedding model based on the Bellman-Ford algorithm, and NodePiece \cite{NodePiece} uses tokenized anchor nodes and relational paths to represent new entities. However, they do not consider learning new knowledge and cannot support new relations.

\subsection{Lifelong Learning}
\label{subsec:lifelong_learning}

Lifelong learning seeks to solve new problems quickly without catastrophically forgetting previously acquired knowledge. 
Lifelong learning models are broadly classified into three categories.
(\romannumeral1) Dynamic architecture models \cite{PNN,CWR} extend the network to learn new tasks and avoid forgetting acquired knowledge. 
(\romannumeral2) Regularization-based models \cite{EWC,SI} capture the importance of model parameters for old tasks and limit the update of important parameters. 
(\romannumeral3) Rehearsal-based models \cite{GEM,EMR} memorize some data from old tasks and replay them when learning new knowledge. 

Few lifelong learning models focus on KG embedding. 
DiCGRL \cite{DiCGRL} is a disentangle-based lifelong graph embedding model. 
It splits node embeddings into different components and replays related historical facts to avoid catastrophic forgetting. 
The work \cite{CKGE} combines class-incremental learning models with TransE~\cite{TransE} for continual KG embedding. 
However, it does not propose a specific lifelong KG embedding model. 

% In addition, there are some studies that focus on continual learning tasks on other graphs. 
% \citet{LLEG} propose a lifelong learning model for evolving graphs with limited labeled data and unseen label classes. 
% TWP \cite{TWP} proposes a topology-aware weight preserving model for the node class incremental task. 
% \citet{FGN} develop a feature graph network to learn growing features. 

% ========================>> Method
\section{Lifelong Knowledge Graph Embedding}
\label{sec:method}

%In this section, we first introduce our problem setting. 
%Then, we present our model, \modelname, in detail. 

\begin{figure*}[!t]
\centering
\includegraphics[width=0.9\textwidth]{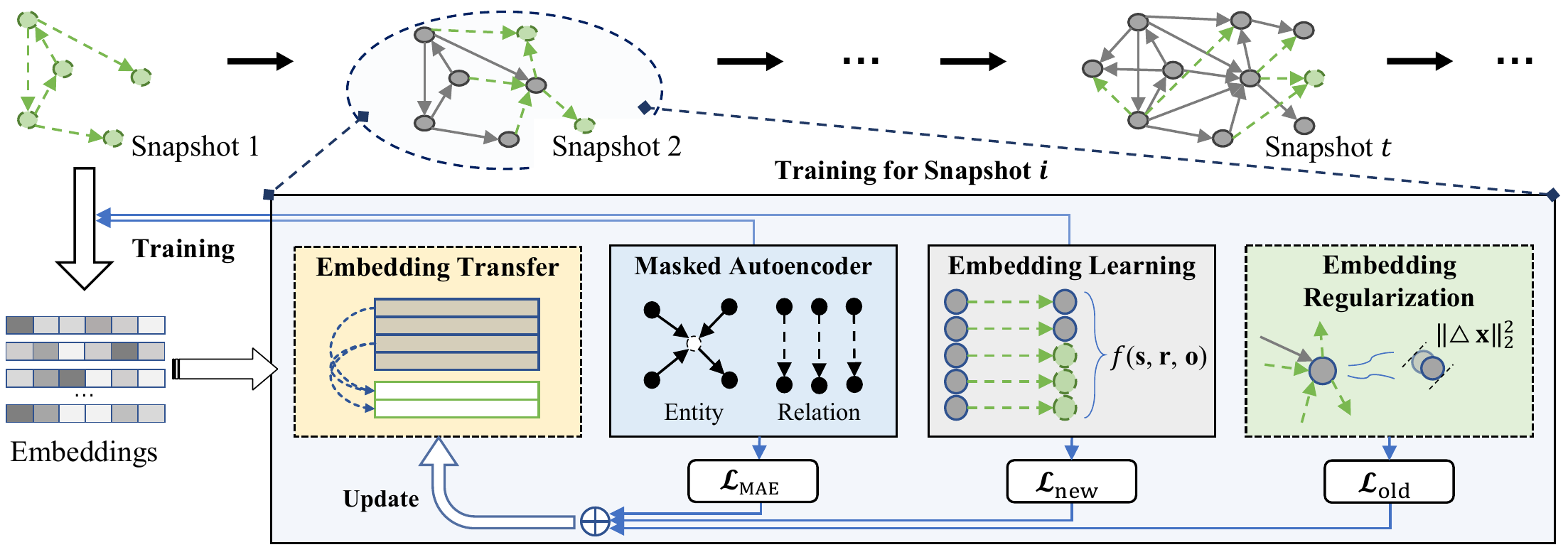}
\caption{Overview of the proposed model for lifelong KG embedding.}
\label{fig:overview}
\centering
\vspace{-5pt}
\end{figure*}

\subsection{Preliminaries}

\textbf{Growing KG.} 
The growth process of a KG yields a snapshot sequence, i.e., $\mathcal{G}=\{\mathcal{S}_1, \mathcal{S}_2, \ldots, \mathcal{S}_t\}$. 
Each snapshot $\mathcal{S}_i$ is defined as a triplet $(\mathcal{T}_i,\mathcal{E}_i, \mathcal{R}_i)$, 
where $\mathcal{T}_i, \mathcal{E}_i$ and $\mathcal{R}_i$ denote the fact, entity and relation sets, respectively.
We have $\mathcal{T}_i \subseteq \mathcal{T}_{i+1}$, $\mathcal{E}_i \subseteq \mathcal{E}_{i+1}$ and $\mathcal{R}_i \subseteq \mathcal{R}_{i+1}$.
We use $\mathcal{T}_{\Delta i}=\mathcal{T}_{i}-\mathcal{T}_{i-1}$, $\mathcal{E}_{\Delta i}=\mathcal{E}_{i}-\mathcal{E}_{i-1}$, and $\mathcal{R}_{\Delta i}=\mathcal{R}_{i}-\mathcal{R}_{i-1}$ to denote the new facts, entities and relations, respectively.
Each fact is in the form of $(s, r, o)\in\mathcal{T}_i$, where $s,o\in\mathcal{E}_i$ are the subject and object entities, respectively, and $r\in\mathcal{R}_i$ is their relation. 

\smallskip
\noindent\textbf{Lifelong KG embedding.} 
KG embedding seeks to encode the symbolic representations of entities and relations into vector space and capture KG semantics using vector operations. 
For a growing KG, a lifelong KG embedding model learns to represent the snapshot sequence $\mathcal{G}=\{\mathcal{S}_1, \mathcal{S}_2, \ldots, \mathcal{S}_t\}$ continually. 
When a new fact set $\mathcal{T}_{\Delta i}$ emerges, the current KG embedding model $\mathcal{M}_{ i-1}$ needs update to fit the new facts and learn embeddings for the new entities $\mathcal{E}_{\Delta i}$ and new relations $\mathcal{R}_{\Delta i}$. 
The resulting model is denoted by $\mathcal{M}_i$.

\smallskip
\noindent\textbf{Lifelong link prediction.} 
The link prediction task asks the KG embedding model to predict the missing subject or object entity in an incomplete fact like $(s, r, ?)$ or $(?, r, o)$.
For each snapshot $\mathcal{S}_i$, the new fact set $\mathcal{T}_{\Delta i}$ is divided into a training set $\mathcal{D}_i$, a validation set $\mathcal{V}_i$ and a test set $\mathcal{Q}_i$. 
In the lifelong setting, the model is required to learn the training data sets, $\mathcal{D}_1,\mathcal{D}_2,\ldots,\mathcal{D}_t$, in turn. 
After finishing the learning on $\mathcal{D}_i$, the model is evaluated on the accumulated test data, which is $\cup_{j=1}^i\mathcal{Q}_j$, to assess the overall learning performance.
After learning $\mathcal{D}_i$, then $\mathcal{D}_i,\mathcal{V}_i$ would no longer be available for the following learning.
Note that the goal of lifelong KG embedding is to improve the overall performance on all snapshots, which requires the KG embedding model to continually learn new knowledge and retain the learned knowledge.

\subsection{Model Overview}
The overview of our model, \modelname, is shown in Figure~\ref{fig:overview}.
It continually learns knowledge over a sequence of KG snapshots without re-training on previously seen data. 
The foundation is a masked KG autoencoder that can reconstruct the entity and relation embeddings from the masked related subgraph.
To enable knowledge transfer from old snapshots to the new one, we propose embedding transfer to inject learned knowledge into unseen entities and relations and then iteratively optimize the model to fit the new facts.
We also propose a lightweight regularization method to retain old knowledge. 

\subsection{Masked Knowledge Graph Autoencoder}
In lifelong KG embedding learning, the new facts, e.g., $\mathcal{T}_{\Delta i}$, bring new entities $\mathcal{E}_{\Delta i}$, and new relations $\mathcal{R}_{\Delta i}$. 
It would also involve some old entities from $\mathcal{E}_{i-1}$, and old relations from $\mathcal{R}_{i-1}$.
Thus, a new entity may be connected with both new and old entities (referred to as new knowledge).
The involved old entities also receive more facts, and therefore their previously learned embeddings (referred to as old knowledge) need to be updated to fit the new facts.
Hence, the base KG embedding model for lifelong learning should be capable of capturing new knowledge as well as updating old knowledge.
To this end, we propose a masked KG autoencoder motivated by the recent success of self-supervised learning~\cite{GraphMAE}.
The key idea is to reconstruct the embedding for an entity or relation based on its masked subgraph, which may include both other new entities and some old entities.
Specifically, we use the first-order subgraph of an entity or relation to reconstruct its embedding $\bar{\mathbf{x}}_i$:
\begin{equation}
    \bar{\mathbf{x}}_i = \text{MAE}\big(\cup_{j=1}^i\mathcal{N}_{j}(x)\big),
\end{equation}
where $x$ denotes either an entity or a relation, and $\mathcal{N}_{j}\subseteq\mathcal{D}_j$ denotes the involved facts of $x$ in the $j$-th snapshot. $\text{MAE}()$ is an encoder to represent the input subgraph.
The objective of our KG encoder is to align the entity or relation embedding with the reconstructed representation as follows:
\begin{equation}
    \mathcal{L}_\text{MAE} = \sum_{e\in\mathcal{E}_i}\Vert\mathbf{e}_i-\bar{\mathbf{e}}_i\Vert_2^2 + \sum_{r\in\mathcal{R}_i}\Vert\mathbf{r}_i-\bar{\mathbf{r}}_i\Vert_2^2.
\end{equation}

The key then becomes how to design an effective and efficient encoder for lifelong learning.
GCN~\cite{Kipf_GCN} and Transformer~\cite{Transformer} are two common choices for the encoder. 
The two encoders both introduce additional model parameters (e.g., the weight matrices).
In our lifelong learning setting, the encoder needs to be updated to fit new facts or subgraphs.
In this case, once the GCN or Transformer is updated, the changed model parameters would affect the embedding generation of \emph{all} old entities (not just the involved old entities in new facts), increasing the risk of catastrophically forgetting previous snapshots. 
To avoid this issue, we use the entity and relation embedding transition functions as encoders, which do not introduce additional parameters.
We borrow the idea of TransE \cite{TransE} and interpret a relation embedding as the translation vector between the subject and object entity embeddings, i.e., $\mathbf{s} + \mathbf{r} \approx \mathbf{o}$,
where $\mathbf{s}, \mathbf{r}, \mathbf{o}$ denote the embeddings of subject entity, relation and object entity, respectively. 
Based on this, we can deduce two transition functions for entity and relation embeddings.
The subject entity of $(s,r,o)$ can be represented by $f_{\rm{sub}}(\mathbf{r}, \mathbf{o})=  \mathbf{o} - \mathbf{r}$,
and the relation embedding is $f_{\rm{rel}}(\mathbf{s}, \mathbf{o})=  \mathbf{o} - \mathbf{s}$.
We can define the encoders as
\begin{align}
\label{eq:auto_ent_lifelong_1}
    \bar{\mathbf{e}}_i &= \frac{\sum_{j=1}^i\sum_{(s, r, o) \in \mathcal{N}_j(e)} f_{\rm{sub}}(\mathbf{r}_i, \mathbf{o}_i)}{\sum_{j=1}^{i} |\mathcal{N}_j(e)|},\\
\label{eq:auto_rel_lifelong_1}
\bar{\mathbf{r}}_i &= \frac{\sum_{j=1}^i\sum_{(s, r, o) \in \mathcal{N}_j(r)} f_{\rm{rel}}(\mathbf{s}_i, \mathbf{o}_i)}{\sum_{j=1}^{i} |\mathcal{N}_j(r)|},
\end{align}
where $\mathbf{s}_i, \mathbf{r}_i, \mathbf{o}_i$ are the embeddings of $s,r,o$ during the training on $\mathcal{D}_i$. 
$\mathcal{N}_j(x)\subseteq\mathcal{D}_j$ is the set of facts containing $x$.

In lifelong learning, the model learns from a snapshot sequence. 
Eqs.~(\ref{eq:auto_ent_lifelong_1}) and (\ref{eq:auto_rel_lifelong_1}) require training samples from the first $i$ snapshots, which are not in line with lifelong learning. 
To reduce the reliance on learned data, we use $\mathbf{e}_{i-1}$ and $\mathbf{r}_{i-1}$ as the approximate average embeddings of $e$ and $r$ in the first $i-1$ snapshots, respectively, and rewrite the encoders as
\begin{equation}
\resizebox{.9\columnwidth}{!}{$
\label{eq:auto_ent_lifelong_2}
\bar{\mathbf{e}}_i \approx \frac{\sum_{j=1}^{i-1}|\mathcal{N}_j(e)| \mathbf{e}_{i-1}+\sum_{(s, r, o) \in \mathcal{N}_i(e)} f_{\rm{sub}}(\mathbf{r}_i, \mathbf{o}_i)}
    {\sum_{j=1}^{i-1}|\mathcal{N}_j(e)|+ |\mathcal{N}_i(e)|},
$}
\end{equation}
\begin{equation}
\resizebox{.9\columnwidth}{!}{$
\label{eq:auto_rel_lifelong_2}
\bar{\mathbf{r}}_i \approx \frac{\sum_{j=1}^{i-1}|\mathcal{N}_j(r)| \mathbf{r}_{i-1}+\sum_{(s, r, o) \in \mathcal{N}_i(r)} f_{\rm{rel}}(\mathbf{s}_i, \mathbf{o}_i)}
    {\sum_{j=1}^{i-1}|\mathcal{N}_j(r)| + |\mathcal{N}_i(r)|}.
$}
\end{equation}

The encoders use the facts involving both old and new entities and relations for embedding reconstruction and they build a bridge for knowledge transfer.

For each snapshot, to learn the knowledge from the new data and update the learned parameters, we leverage TransE \cite{TransE} to train the embedding model:
\begin{equation}
\resizebox{.9\columnwidth}{!}{$
    \mathcal{L}_\text{new} = \sum\limits_{(s, r, o)\in\mathcal{D}_i} \max\big(0,\gamma+f(\mathbf{s}, \mathbf{r}, \mathbf{o})-f(\mathbf{s}', \mathbf{r}, \mathbf{o}')\big),
$}
\end{equation}
where $\gamma$ is the margin. $(\mathbf{s}', \mathbf{r}, \mathbf{o}')$ is the embedding of a negative fact. 
For each positive fact, we randomly replace the subject or object entity with a random entity $e'\in\mathcal{E}_i$.

% To enhance the knowledge co-transfer between related tasks, we use the cross entropy loss function for learning to reconstruct the embeddings of entities and relations:
% \begin{equation}
%     \mathcal{L}_{MAE} = \sum_{e\in\mathcal{E}_i}\Vert\mathbf{e}_i-\bar{\mathbf{e}}_i\Vert_2^2 + \sum_{r\in\mathcal{R}_i}\Vert\mathbf{r}_i-\bar{\mathbf{r}}_i\Vert_2^2.
% \end{equation}

\subsection{Embedding Transfer}
%To accelerate the learning of new snapshots, our first challenge is how to learn the embeddings of unseen entities and relations based on the learned knowledge. 
During the lifecycle of a growing KG, there are abundant unseen entities and some unseen relations emerge with the new facts. 
Learning effective embeddings for them is an essential aspect of lifelong KG embedding learning.
However, these unseen ones are not included in any learned snapshots, so only inheriting the learned parameters cannot transfer the acquired knowledge to their embeddings.
To avoid learning from scratch, we propose embedding transfer that seeks to leverage the learned embeddings to help represent unseen entities and relations.
Specifically, we initialize the embeddings of each unseen entity by aggregating its facts:
\begin{equation}
    \mathbf{e}_{i} = \frac{1}{|\mathcal{N}_i(e)|}\sum_{(e, r, o) \in \mathcal{N}_i(e)} f_{\rm{sub}}(\mathbf{r}_{i-1}, \mathbf{o}_{i-1}) ,
\end{equation}
where $\mathcal{N}_i(e)\subseteq\mathcal{D}_i$ is the set of facts containing $e$. 
% In this paper, we adopt a widely used translation-based score function \cite{TransE} with the L1-norm: $f(\mathbf{s}, \mathbf{r}, \mathbf{o}) = \Vert\mathbf{s} + \mathbf{r} - \mathbf{o}\Vert$, where $\mathbf{s}, \mathbf{r}, \mathbf{o}$ are the embeddings of subject entity, relation and object entity. The inverse score functions for subject entity is $f_{\rm{sub}}(\mathbf{r}, \mathbf{o})= 0 - \mathbf{r} + \mathbf{o}$.
For the new entities that do not have common facts involving existing entities, we randomly initialize their embeddings.
We also use this strategy to initialize the embeddings of unseen relations:
\begin{equation}
    \mathbf{r}_i = \frac{1}{|\mathcal{N}_i(r)|}\sum_{(s, r, o) \in \mathcal{N}_i(r)} f_{\rm{rel}}(\mathbf{s}_{i-1}, \mathbf{o}_{i-1}),
\end{equation}
where $\mathcal{N}_i(r)\subseteq\mathcal{D}_i$ is the set of facts containing $r$.

\subsection{Embedding Regularization}
Learning new snapshots is likely to overwrite the learned knowledge from old snapshots. 
To avoid catastrophic forgetting, some regularization methods \cite{EWC,GEM} constrain the updates of parameters that are important to old tasks.
The loss function of regularization methods is
\begin{equation}
\resizebox{.88\columnwidth}{!}{$
    \mathcal{L}_\text{old} = \sum\limits_{e\in\mathcal{E}_{i-1}}\omega(e)\Vert\mathbf{e}_i-\mathbf{e}_{i-1}\Vert_2^2 + \sum\limits_{r\in\mathcal{R}_{i-1}}\omega(r)\Vert\mathbf{r}_i-\mathbf{r}_{i-1}\Vert_2^2,
    $}
\end{equation}
where $\omega(x)$ is the regularization weight for $x$.

Conventional regularization-based methods for classification tasks such as \cite{EWC} model the importance of each parameter at a high cost based on the gradient or parameter change during training. 
This problem is even more severe for KG embedding models that have a large number of embedding parameters (i.e. entity and relation embeddings).
To resolve this problem, we propose a lightweight embedding regularization method, which calculates the regularization weight of each entity or relation by the number of new and old facts containing it:
\begin{equation}
    \omega(x) = 1 - \frac{|\mathcal{N}_{i}(x)|}{\sum_{j=1}^i|\mathcal{N}_{j}(x)|}.
\end{equation}

% Note that some models contain parameters in addition to entity and relation embeddings. The $\mathcal{N}_i(x)$ of specific entity- or relation-related parameters (such as the specific relation-related message function parameters in R-GCN) is same as the entity or relation. And the $\mathcal{N}_i(x)$ for other parameters is set as the training set $\mathcal{D}_i$, since they are directly related with all facts in $\mathcal{D}_i$. 
As a lightweight technique, it only keeps the total number of involved trained facts for each entity or relation and only updates regularization weights once per snapshot.

\subsection{Overall Learning Objective}
To learn new knowledge while retaining acquired knowledge, the overall lifelong learning objective $\mathcal{L}$ is defined as follows:
\begin{equation}
    \mathcal{L} = \mathcal{L}_\text{new} + \alpha\,\mathcal{L}_\text{old} + \beta\,\mathcal{L}_\text{MAE},
\end{equation}
where $\alpha,\beta$ are hyperparameters for balancing the objectives.

\subsection{Complexity Analysis}
Compared with fine-tuning, the proposed model requires few extra resources. 
It does not increase the size of training samples like the rehearsal models \cite{EMR, GEM, DiCGRL}. 
In addition to fine-tuning, the proposed model calculates the loss of masked autoencoder and embedding regularization. 
The additional time complexity in each iteration is $O(|\mathcal{E}|+|\mathcal{R}|+|\mathcal{D}|)$. 
In practice, we find that the loss of autoencoder can accelerate learning, and its time consumption is close to that of fine-tuning. 
The space complexity of fine-tuning is $O((|\mathcal{E}|+|\mathcal{R}|)\times d)$, and the space complexity of the proposed model is $O((|\mathcal{E}|+|\mathcal{R}|)\times(d+1))$, where $d$ is the dimension of embeddings.

% =======================> datasets
\begin{table*}
\centering
\setlength\tabcolsep{4.4pt}
\resizebox{\linewidth}{!}{
    \begin{tabular}{lcccccccccccccccccccc}
    \toprule
    \multirow{2}{*}{Datasets} & \multicolumn{3}{c}{Snapshot 1} & \multicolumn{3}{c}{Snapshot 2} &    \multicolumn{3}{c}{Snapshot 3} & \multicolumn{3}{c}{Snapshot 4} & \multicolumn{3}{c}{Snapshot 5} \\ 
    \cmidrule(lr){2-4} \cmidrule(lr){5-7} \cmidrule(lr){8-10} \cmidrule(lr){11-13} \cmidrule(lr){14-16} & $|\mathcal{T}_{\Delta 1}|$ & $|\mathcal{E}_1|$ & $|\mathcal{R}_1|$ & $|\mathcal{T}_{\Delta 2}|$ & $|\mathcal{E}_2|$ & $|\mathcal{R}_2|$ & $|\mathcal{T}_{\Delta 3}|$ & $|\mathcal{E}_3|$ & $|\mathcal{R}_3|$ & $|\mathcal{T}_{\Delta 4}|$ & $|\mathcal{E}_4|$ & $|\mathcal{R}_4|$ & $|\mathcal{T}_{\Delta 5}|$ & $|\mathcal{E}_5|$ & $|\mathcal{R}_5|$ \\
    \midrule
    \textsc{Entity} & 46,388 & \ \ 2,909 & 233 & 72,111 & \ \ 5,817 & 236 & 73,785 & \ \ 8,275 & 236 & \ \ 70,506 & 11,633 & 237 & 47,326 & 14,541 & 237  \\
    \textsc{Relation} & 98,819 & 11,560 & \ \ 48 & 93,535 & 13,343 & \ \ 96 & 66,136 & 13,754 & 143 & \ \ 30,032 & 14,387 & 190 & 21,594 & 14,541 & 237  \\
    \textsc{Fact} & 62,024 & 10,513 & 237 & 62,023 & 12,779 & 237 & 62,023 & 13,586 & 237 & \ \ 62,023 & 13,894 & 237 & 62,023 & 14,541 & 237  \\
    \textsc{Hybrid} & 57,561 & \ \ 8,628 & \ \ 86 & 20,873 & 10,040 & 102 & 88,017 & 12,779 & 151 & 103,339 & 14,393 & 209 & 40,326 & 14,541 & 237 \\
    \bottomrule
\end{tabular}}
\caption{Statistical data of the four constructed growing KG datasets. 
For the $i$-th snapshot, $\mathcal{T}_{\Delta i}$ denotes the set of new facts in this snapshot, and $\mathcal{E}_i, \mathcal{R}_{i}$ denote the sets of cumulative entities and relations in the first $i$ snapshots, respectively.}
\label{tab:datasets}
\end{table*}

% ========================>> Datasets
\section{Dataset Construction}
\label{sec:datasets}

To simulate a variety of aspects of KG growth, we create four datasets based on FB15K-237 \cite{FB237}, which are entity-centric, relation-centric, fact-centric, and hybrid.
We denote them by \textsc{Entity}, \textsc{Relation}, \textsc{Fact} and \textsc{Hybrid}, respectively. 
Given a KG $\mathcal{G}=\{\mathcal{E}, \mathcal{R}, \mathcal{T}\}$, we construct five snapshots with the following steps:
\begin{enumerate}
\item \textbf{Seeding.} 
We randomly sample 10 facts from $\mathcal{T}$ and add them into $\mathcal{T}_1$ for initialization. 
The entities and relations in the 10 facts form the initial $\mathcal{E}_1$ and $\mathcal{R}_1$, respectively.

\item \textbf{Expanding.} 
To build \textsc{Entity}, \textsc{Relation} and \textsc{Fact}, we iteratively sample a fact containing at least one seen entity in $\mathcal{E}_i$, add it into $\mathcal{T}_i$, and extract the unseen entity and relation from it to expand $\mathcal{E}_i$ and $\mathcal{R}_i$. 
For \textsc{Entity}, once $|\mathcal{E}_i| \geq \frac{i+1}{5}|\mathcal{E}|$, we add all new facts $\big\{(s, r, o)\,|\,s\in\mathcal{E}_i\wedge o\in\mathcal{E}_i \big\}$ into $\mathcal{T}_i$ and start building the next snapshot. 
In the same way, we construct \textsc{Relation} and \textsc{Fact}. 
As for \textsc{Hybrid}, we uniformly sample an entity, relation or fact without replacement from $\mathcal{U}=\mathcal{E}\cup\mathcal{R}\cup\mathcal{T}$ to join $\mathcal{E}_i$, $\mathcal{R}_i$ and $\mathcal{T}_i$. 
Note that when the sampled fact contains an unseen entity or relation, we re-sample a fact that only contains seen entities and relations to replace it.
After each iteration, we terminate the expansion of this snapshot with a probability $\frac{5}{|\mathcal{U}|}$. 
Consequently, the expansion of \textsc{Hybrid} is uneven, making it more realistic and challenging.
For all datasets, we take the whole KG as the last snapshot, i.e., $\mathcal{T}_5=\mathcal{T}$, and $\mathcal{E}_5=\mathcal{E}, \mathcal{R}_5=\mathcal{R}$.

\item \textbf{Dividing.} 
For each snapshot, we randomly divide the new fact set $\mathcal{T}_{\Delta i}$ into a training set $\mathcal{D}_i$, a validation set $\mathcal{V}_i$ and a test set $\mathcal{Q}_i$ by a split ratio of 3:1:1. 
%We ensure that $\mathcal{V}_i$ and $\mathcal{Q}_i$ only contain entities and relations in $\cup_{j=1}^i \mathcal{D}_j$.
\end{enumerate}

The statistics of the four datasets are presented in Table~\ref{tab:datasets}. 

% ========================>> Experiments
\section{Experimental Results}
\label{sec:experiments}

We conduct experiments regarding link prediction accuracy, knowledge transfer capability, and learning efficiency to validate the proposed model, LKGE.
The datasets and source code are available at \url{https://github.com/nju-websoft/LKGE}.

\subsection{Experiment Settings}
\noindent\textbf{Competitors.} We compare our model with 12 competitors, including 
(\romannumeral1) three baseline models: snapshot only, re-training, and fine-tuning; 
(\romannumeral2) two inductive models: MEAN \cite{MEAN}, and LAN \cite{LAN}; 
(\romannumeral3) two dynamic architecture models: PNN \cite{PNN}, and CWR \cite{CWR};
(\romannumeral4) two regularization-based models: SI \cite{SI}, and EWC \cite{EWC}; 
and (\romannumeral5) three rehearsal-based models: GEM \cite{GEM}, EMR \cite{EMR}, and DiCGRL \cite{DiCGRL}.

\smallskip
\noindent\textbf{Evaluation metrics.} Following the convention, we conduct the experiments on link prediction. 
Given a snapshot $\mathcal{S}_i$, for each test fact $(s, r, o)\in\mathcal{Q}_i$, we construct two queries $(s, r, ?)$ and $(?, r, t)$. 
When evaluating on $\mathcal{Q}_i$, we set all seen entities in $\mathcal{E}_i$ as candidate entities. 
We select seven metrics to evaluate all models, including 
(\romannumeral1) Four metrics on link prediction accuracy: mean reciprocal rank (MRR) and Hits@$k$ ($k=1,3,10$, and H@$k$ for shot). 
We conduct the model $\mathcal{M}_5$ trained on the last snapshot to evaluate on the union of the test sets in all snapshots.
(\romannumeral2) Two metrics on knowledge transfer capability: forward transfer (FWT) and backward transfer (BWT) \cite{GEM}.
FWT is the influence of learning a task to the performance on the future tasks, while BWT is the influence of learning to the previous tasks: 
\begin{align}
\resizebox{.88\columnwidth}{!}{$
\text{FWT} = \frac{1}{n-1}\sum\limits_{i=2}^{n} h_{i-1,i},\ \text{BWT} = \frac{1}{n-1}\sum\limits_{i=1}^{n-1} (h_{n,i} - h_{i,i}),
$}
\end{align}
where $n$ is the number of snapshots, $h_{i,j}$ is the MRR scores on $\mathcal{Q}_j$ after training the model $\mathcal{M}_i$ on the $i$-th snapshot.
Higher scores indicate better performance.
(\romannumeral3) Time cost: The cumulative time cost of the learning on each snapshot.

\begin{table*}[!t]
\centering
\setlength\tabcolsep{5pt}
\resizebox{1.0\textwidth}{!}{\Large
\begin{tabular}{lcccccccccccccccc}
\toprule
\multirow{2}{*}{Models} & \multicolumn{4}{c}{\textsc{Entity}} & \multicolumn{4}{c}{\textsc{Relation}} & \multicolumn{4}{c}{\textsc{Fact}} & \multicolumn{4}{c}{\textsc{Hybrid}} \\
\cmidrule(lr){2-5} \cmidrule(lr){6-9} \cmidrule(lr){10-13} \cmidrule(lr){14-17} & MRR & H@1 & H@3 & H@10 & MRR & H@1 & H@3 & H@10 & MRR & H@1 & H@3 & H@10 & MRR & H@1 & H@3 & H@10 \\
\midrule
Snapshot	& 0.084 	& 0.028 	& 0.107 	& 0.193 	& 0.021 	& 0.010 	& 0.023 	& 0.043 	& 0.082 	& 0.030 	& 0.095 	& 0.191 	& 0.036 	& 0.015 	& 0.043 	& 0.077 	\\
Re-train	& 0.236 	& 0.137 	& 0.274 	& 0.433 	& 0.219 	& 0.128 	& 0.250 	& 0.403 	& 0.206 	& 0.118 	& 0.232 	& 0.385 	& 0.227 	& 0.134 	& 0.260 	& 0.413 	\\
Fine-tune	& 0.165 	& 0.085 	& 0.188 	& 0.321 	& 0.093 	& 0.039 	& 0.106 	& 0.195 	& 0.172 	& 0.090 	& 0.193 	& 0.339 	& 0.135 	& 0.069 	& 0.151 	& 0.262 	\\
\midrule
MEAN	& 0.117 	& 0.068 	& 0.123 	& 0.212 	& 0.039 	& 0.024 	& 0.040 	& 0.067 	& 0.084 	& 0.051 	& 0.088 	& 0.146 	& 0.046 	& 0.029 	& 0.049 	& 0.080 	\\
LAN	& 0.141 	& 0.082 	& 0.149 	& 0.256 	& 0.052 	& 0.033 	& 0.052 	& 0.092 	& 0.106 	& 0.056 	& 0.113 	& 0.200 	& 0.059 	& 0.032 	& 0.062 	& 0.113 	\\
\midrule
PNN	& 0.229 	& 0.130 	& 0.265 	& \textbf{0.425} 	& 0.167 	& 0.096 	& 0.191 	& 0.305 	& 0.157 	& 0.084 	& 0.188 	& 0.290 	& 0.185 	& 0.101 	& 0.216 	& 0.349 	\\
CWR	& 0.088 	& 0.028 	& 0.114 	& 0.202 	& 0.021 	& 0.010 	& 0.024 	& 0.043 	& 0.083 	& 0.030 	& 0.095 	& 0.192 	& 0.037 	& 0.015 	& 0.044 	& 0.077 	\\
\midrule
SI	& 0.154 	& 0.072 	& 0.179 	& 0.311 	& 0.113 	& 0.055 	& 0.131 	& 0.224 	& 0.172 	& 0.088 	& 0.194 	& 0.343 	& 0.111 	& 0.049 	& 0.126 	& 0.229 	\\
EWC	& 0.229 	& 0.130 	& 0.264 	& 0.423 	& 0.165 	& 0.093 	& 0.190 	& 0.306 	& 0.201 	& 0.113 	& 0.229 	& 0.382 	& 0.186 	& 0.102 	& 0.214 	& 0.350 	\\
\midrule
GEM	& 0.165 	& 0.085 	& 0.188 	& 0.321 	& 0.093 	& 0.040 	& 0.106 	& 0.196 	& 0.175 	& 0.092 	& 0.196 	& 0.345 	& 0.136 	& 0.070 	& 0.152 	& 0.263 	\\
EMR	& 0.171 	& 0.090 	& 0.195 	& 0.330 	& 0.111 	& 0.052 	& 0.126 	& 0.225 	& 0.171 	& 0.090 	& 0.191 	& 0.337 	& 0.141 	& 0.073 	& 0.157 	& 0.267 	\\
DiCGRL	& 0.107 	& 0.057 	& 0.110 	& 0.211 	& 0.133 	& 0.079 	& 0.147 	& 0.241 	& 0.162 	& 0.084 	& 0.189 	& 0.320 	& 0.149 	& 0.083 	& 0.168 	& 0.277 	\\
\midrule
LKGE	& \textbf{0.234} 	& \textbf{0.136} 	& \textbf{0.269} 	& \textbf{0.425} 	& \textbf{0.192} 	& \textbf{0.106} 	& \textbf{0.219} 	& \textbf{0.366} 	& \textbf{0.210} 	& \textbf{0.122} 	& \textbf{0.238} 	& \textbf{0.387} 	& \textbf{0.207} 	& \textbf{0.121} 	& \textbf{0.235} 	& \textbf{0.379} 	\\
\bottomrule
\end{tabular}}
\caption{Result comparison of link prediction on the union of the test sets in all snapshots.}
\label{tab:main_results}
\vspace{-5pt}
\end{table*}

\smallskip
\noindent\textbf{Implementation details.}
We use TransE \cite{TransE} as the base model and modify the competitors to do our task:
\begin{itemize}
    \item \textit{Snapshot only}. For the $i$-th snapshot, we reinitialize and train a model only on the training set $\mathcal{D}_i$.
    
    \item \textit{Re-training}. For the $i$-th snapshot, we reinitialize and train a model on the accumulated training data $\cup_{j=1}^i \mathcal{D}_j$.
    
    \item \textit{Fine-tuning}. For the $i$-th snapshot, the model inherits the learned parameters of the model trained on the previous snapshots, and we incrementally train it on $\mathcal{D}_i$.
    
    \item \textit{Inductive} models. We train each model on the first snapshot and obtain the embeddings of unseen entities in the following snapshots by neighborhood aggregation.
    
    \item \textit{Dynamic architecture} models. For PNN, 
    following the implementation of \cite{CKGE}, 
    % when learning on a new snapshot,
    we freeze the parameters learned on previous snapshots and update new parameters. For CWR, after training on $\mathcal{D}_1$, we replicate a model as the consolidated model. For the following $i$-th snapshot, we reinitialize and train a temporal model on $\mathcal{D}_i$, and merge the temporal model into the consolidated model by copying new parameters or averaging old ones.
    
    \item \textit{Regularization} models. Since the base model parameters increase with the emergence of unseen entities and relations, we only use the parameters learned from the previous snapshot to calculate the regularization loss.
    
    \item \textit{Rehearsal} models. We store 5,000 training facts from previous snapshots and add them to the current training set of the $i$-th snapshot. After the learning, we randomly replace half of these facts with those in $\mathcal{D}_i$.
\end{itemize}

% The details of hyper-parameter selection can be found in \url{https://arxiv.org/pdf/2211.15845.pdf}.

For a fair comparison, we first tune the hyperparameters of the base model using grid-search: learning rate in \{0.0005, 0.0001, 0.001\}, batch size in \{1024, 2048\}, embedding dimension in \{100, 200\}.
Then, we use the same base model for \modelname and all competitors, and tune other hyperparameters.
For the regularization models, the $\alpha$ of regularization loss is in \{0.01, 0.1, 1.0\}. 
For our model, the $\beta$ of MAE loss is in \{0.01, 0.1, 1.0\}.
For all competitors, we use Adam optimizer and set the patience of early stopping to 3.
%All experiments are conducted with two NVIDIA RTX 3090 GPUs, two Intel Xeon Gold 5122 CPUs, and 384GB RAM.

% =========>> Main Results

\subsection{Link Prediction Accuracy}

In Table~\ref{tab:main_results}, we run 5-seeds experiments for all models on our datasets and report the means.  
The results show that:
(\romannumeral1) our model consistently achieves the best performance across all datasets. Some results of our model in \textsc{Fact} even outperforms re-training. 
This is because our masked KG autoencoder effectively improves information propagation based on both old and new embeddings, and the embedding regularization avoids catastrophic forgetting. 
Most competitors only work well on \textsc{Entity}, while our model shows stable and promising results on all these datasets.
(\romannumeral2) Re-training is far superior to most baseline models on \textsc{Relation} and \textsc{Hybrid}, while the gaps on \textsc{Entity} and \textsc{Fact} are small. 
This is because the KG embedding model learns two aspects of knowledge: relational patterns and entity embeddings. 
In \textsc{Entity} and \textsc{Fact}, the relational patterns are stable, while in \textsc{Relation} and \textsc{Hybrid}, their relational patterns are constantly changing due to unseen relations. 
These phenomena illustrate that the variation of relational patterns is more challenging for lifelong KG embedding.
(\romannumeral3) The inductive models are only trained on the first snapshot, and cannot transfer knowledge to unseen relations. 
So, their results are lower than other models, especially on \textsc{Relation} and \textsc{Hybrid} with many unseen relations. 
(\romannumeral4) Since the learned parameters are not updated, PNN preserves the learned knowledge well. 
But on \textsc{Fact}, due to a few unseen entities, it lacks new learnable parameters, and the performance is not well. 
CWR averages the old and new model parameters, which does not work well on the embedding learning task. 
(\romannumeral5) EWC performs well because it can model the importance of each parameter using Fisher information matrices to the learned snapshots. 
(\romannumeral6) Unlike the classification tasks, most parameters of KG embedding models correspond to specific entities, so the training data cannot be divided into a few types, and we cannot use 5,000 old samples to replay the learned facts for all entities. 
The performance of GEM, EMR and DiCGRL is limited.

\begin{figure}[t]
\centering
\includegraphics[width=.89\columnwidth]{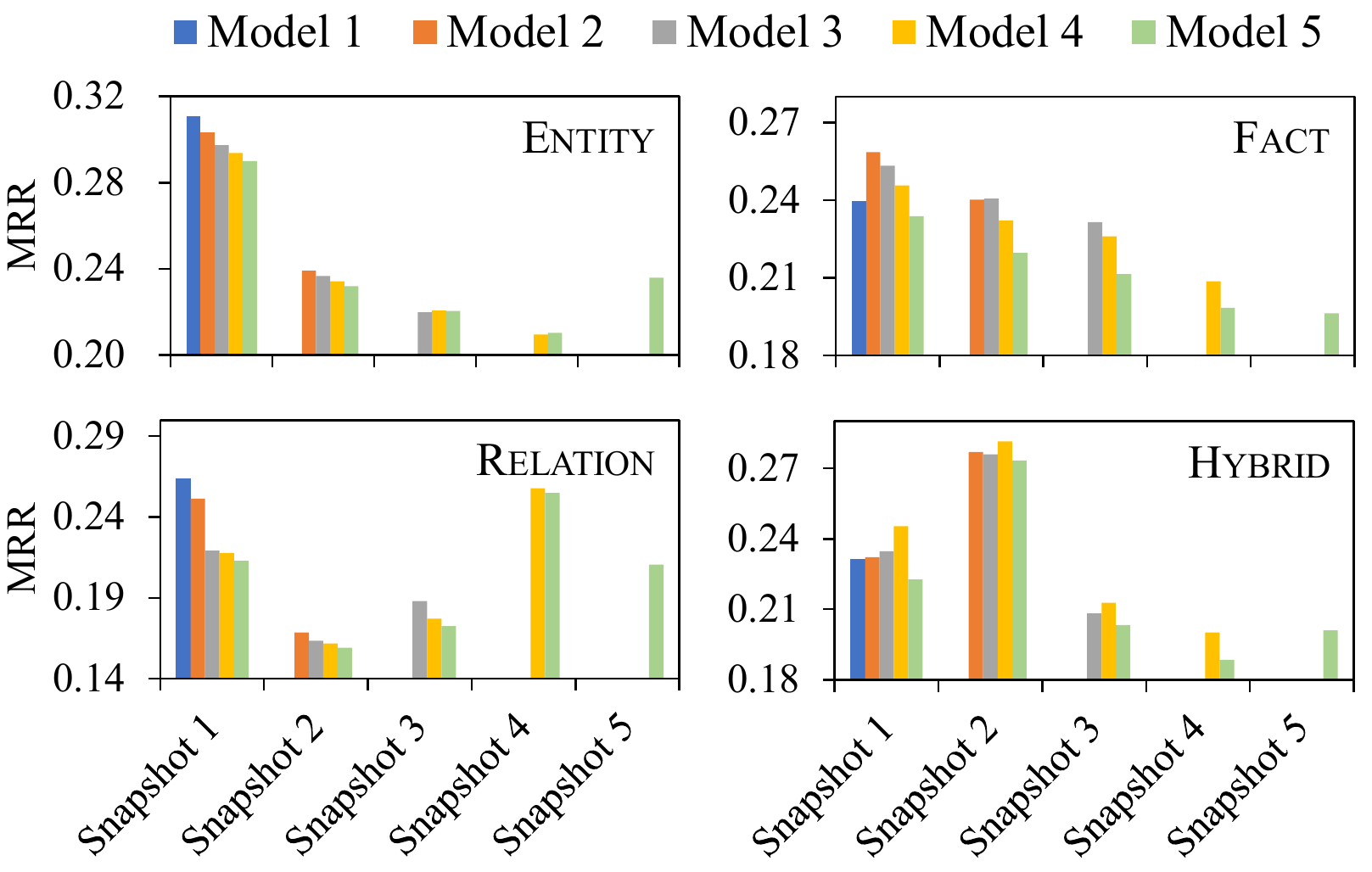}
\caption{MRR changes. $\mathcal{M}_i$ is trained for the $i$-th snapshot and evaluated using the test data of previous snapshots 1 to $i$.}
\label{fig:mrr}
\vspace{-5pt}
\end{figure}

% \begin{table*}[!t]
% \centering
% \setlength\tabcolsep{3.05pt}
% \resizebox{1.0\textwidth}{!}{
% \begin{tabular}{lcccccccccccccccc}
% \toprule
% \multirow{2}{*}{Variants} & \multicolumn{4}{c}{\textsc{Entity}} & \multicolumn{4}{c}{\textsc{Relation}} & \multicolumn{4}{c}{\textsc{Fact}} & \multicolumn{4}{c}{\textsc{Hybrid}} \\
% \cmidrule(lr){2-5} \cmidrule(lr){6-9} \cmidrule(lr){10-13} \cmidrule(lr){14-17} & MRR & H@1 & H@3 & H@10 & MRR & H@1 & H@3 & H@10 & MRR & H@1 & H@3 & H@10 & MRR & H@1 & H@3 & H@10 \\
% \midrule
% LKGE (full) & .234 	& .136 	& .269 	& .425 	& .192 	& .106 	& .219 	& .366 	& .210 	& .122 	& .238 	& .387 	& .207 	& .121 	& .235 	& .379 	\\
% $-$ fine-tuning	& .123	& .068	& .136	& .225	& .126	& .073	& .146	& .231	& .154	& .091	& .176	& .269	& .109	& .060	& .126	& .201	\\
% $-$ autoencoder	& .222	& .124	& .255	& .415	& .185	& .100	& .212	& .355	& .191	& .105	& .215	& .369	& .198	& .111	& .227	& .367	\\
% $-$ embedding transfer	& .240	& .141	& .275	& .433	& .174	& .091	& .201	& .339	& .210	& .123	& .237	& .390	& .200	& .112	& .229	& .372	\\
% $-$ regularization	& .166	& .089	& .184	& .316	& .040	& .014	& .049	& .089	& .175	& .095	& .195	& .338	& .154	& .079	& .171	& .300	\\
% \bottomrule
% \end{tabular}}
% \caption{Ablation results of link prediction on the union of the test sets in all snapshots.}
% \label{tab:ablation}
% \vspace{-5pt}
% \end{table*}

\begin{table*}[!t]
\centering
\setlength\tabcolsep{2.7pt}
\resizebox{1.0\textwidth}{!}{
\begin{tabular}{lcccccccccccccccc}
\toprule
\multirow{2}{*}{Variants} & \multicolumn{4}{c}{\textsc{Entity}} & \multicolumn{4}{c}{\textsc{Relation}} & \multicolumn{4}{c}{\textsc{Fact}} & \multicolumn{4}{c}{\textsc{Hybrid}} \\
\cmidrule(lr){2-5} \cmidrule(lr){6-9} \cmidrule(lr){10-13} \cmidrule(lr){14-17} & MRR & H@1 & H@3 & H@10 & MRR & H@1 & H@3 & H@10 & MRR & H@1 & H@3 & H@10 & MRR & H@1 & H@3 & H@10 \\
\midrule
LKGE (full) & 0.234 	& 0.136 	& 0.269 	& 0.425 	& 0.192 	& 0.106 	& 0.219 	& 0.366 	& 0.210 	& 0.122 	& 0.238 	& 0.387 	& 0.207 	& 0.121 	& 0.235 	& 0.379 	\\
$-$ fine-tuning	& 0.123	& 0.068	& 0.136	& 0.225	& 0.126	& 0.073	& 0.146	& 0.231	& 0.154	& 0.091	& 0.176	& 0.269	& 0.109	& 0.060	& 0.126	& 0.201	\\
$-$ autoencoder	& 0.222	& 0.124	& 0.255	& 0.415	& 0.185	& 0.100	& 0.212	& 0.355	& 0.191	& 0.105	& 0.215	& 0.369	& 0.198	& 0.111	& 0.227	& 0.367	\\
$-$ embedding transfer	& 0.240	& 0.141	& 0.275	& 0.433	& 0.174	& 0.091	& 0.201	& 0.339	& 0.210	& 0.123	& 0.237	& 0.390	& 0.200	& 0.112	& 0.229	& 0.372	\\
$-$ regularization	& 0.166	& 0.089	& 0.184	& 0.316	& 0.040	& 0.014	& 0.049	& 0.089	& 0.175	& 0.095	& 0.195	& 0.338	& 0.154	& 0.079	& 0.171	& 0.300	\\
\bottomrule
\end{tabular}}
\caption{Ablation results of link prediction on the union of the test sets in all snapshots.}
\label{tab:ablation}
\vspace{-5pt}
\end{table*}

To show the performance evolution of \modelname during the learning process, we evaluate the model $\mathcal{M}_i$ trained for the $i$-th snapshot using the test data from previous snapshots.
The MRR results are in Figure~\ref{fig:mrr}. 
\modelname can maintain the learned knowledge during lifelong learning. 
On some snapshots like \textsc{Entity} Snapshot 3, the knowledge update improves the performance on old test data,
which shows that old knowledge update has the potential for backward knowledge transfer.

\subsection{Knowledge Transfer Capability}

\begin{figure}[t]
\includegraphics[width=\columnwidth]{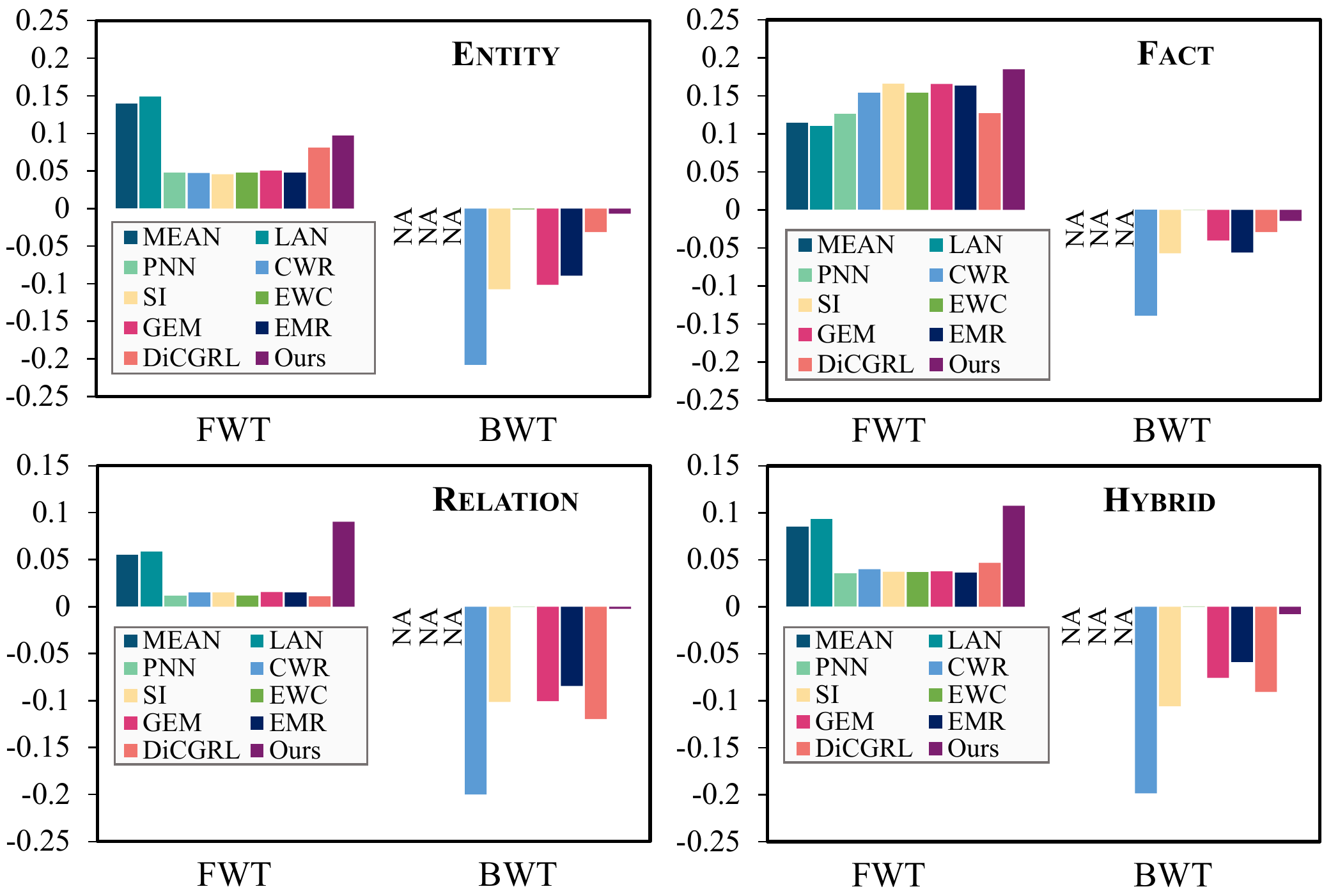}
\caption{Forward transfer and backward transfer of MRR.}
\label{fig:knowledge_transfer}
\vspace{-5pt}
\end{figure}

To evaluate the knowledge transfer and retention capability of all models, we report the FWT and BWT of MRR results in Figure~\ref{fig:knowledge_transfer}. 
Because of the embedding transfer, the FWT of \modelname is higher than all lifelong learning competitors. 
Even on \textsc{Relation} and \textsc{Hybrid} where the KG schema changes, \modelname still keeps the FWT capability well. 
MEAN and LAN are designed to transfer knowledge forward to embed new entities.  
So, they work well on \textsc{Entity}.
However, their FWT capability is limited on other datasets since they cannot update the old embeddings to adapt to new snapshots.

BWT is usually negative due to the overwriting of learned knowledge. 
PNN, MEAN, and LAN do not update old parameters. Their BWT scores are ``NA''. 
The poor scores of CWR show the harmful effects of the average operation.
The scores of rehearsal models are also not good as they cannot store enough facts. 
\modelname gets good BWT scores as the embedding regularization can well maintain the learned knowledge.
% balance the learning of new and the update of old embeddings.

\subsection{Learning Efficiency}
\begin{figure}[t]
\includegraphics[width=\columnwidth]{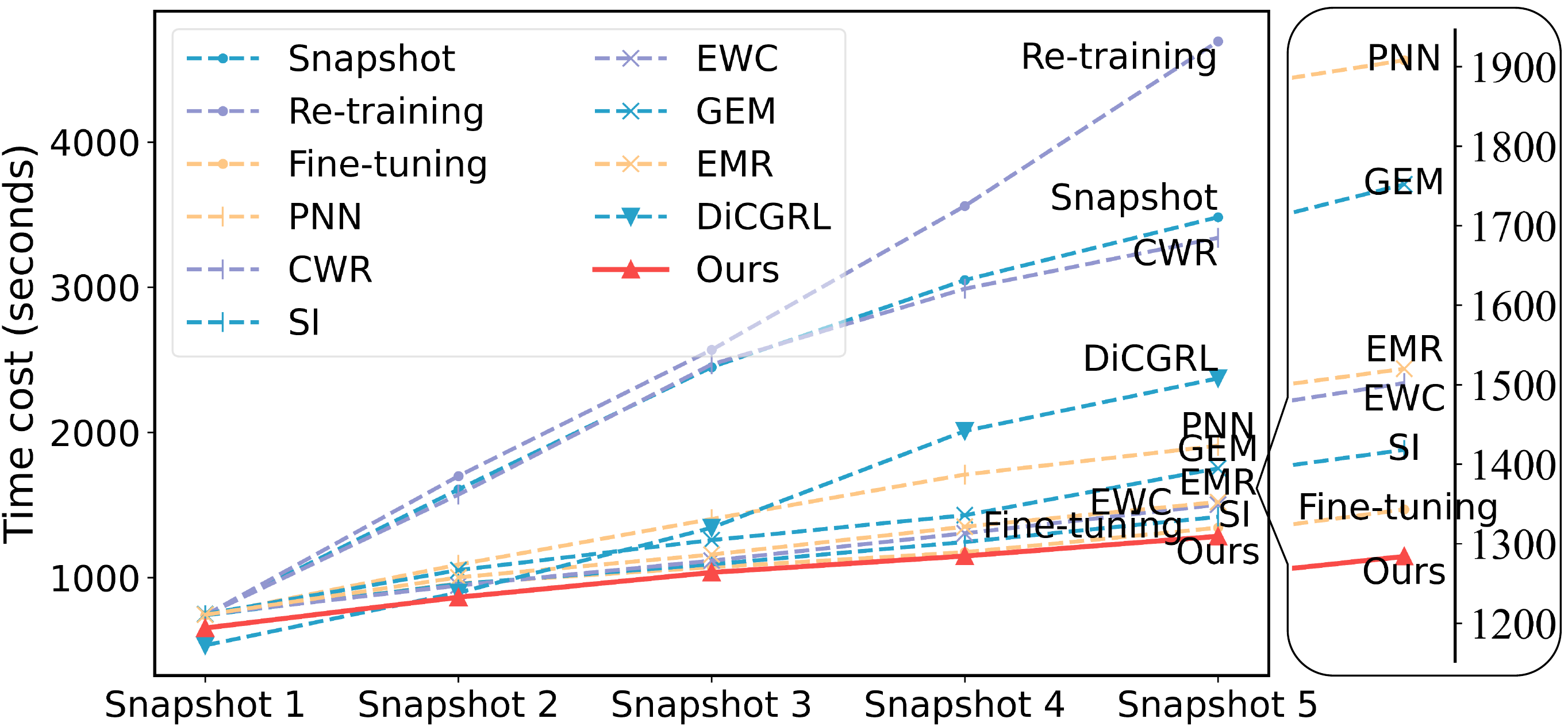}
\caption{Cumulative time cost on \textsc{Fact}.}
\label{fig:time_cost}
\vspace{-5pt}
\end{figure}

% Time & Space
We show the training time on \textsc{Fact} as all snapshots of \textsc{Fact} have the same training set size.
Figure~\ref{fig:time_cost} shows the results.
Unsurprisingly, re-training is most time-consuming. 
Snapshot is also costly because it cannot inherit knowledge from previous snapshots. 
By contrast, our model is most efficient, and its advantage is more significant in the final snapshot. 
This is because the embedding transfer can use the learned knowledge to accelerate the learning of new facts.

% =========>> Ablation Study

\subsection{Ablation Study}
\label{subsec:ablation}
We conduct an ablation study by designing four variants of \modelname: ``w/o fine-tuning'', ``w/o autoencoder'', ``w/o embedding transfer'' and ``w/o regularization''. 
The ``w/o fine-tuning'' variant is trained on $\mathcal{D}_1$ and performs the embedding transfer on other $\mathcal{D}_i$. 
The latter three variants disable the specific components in \modelname. 
The results are shown in Table~\ref{tab:ablation}. 
We see that 
(\romannumeral1) although fine-tuning is disabled, ``w/o fine-tuning'' can still perform well with only the knowledge from the first snapshot, showing that embedding transfer can effectively transfer learned knowledge to unseen entities and relations. 
(\romannumeral2) Both ``w/o autoencoder'' and ``w/o regularization'' significantly drop, showing the effects of masked KG autoencoder and knowledge retention.
(\romannumeral3) Embedding transfer enables the model to be trained at a starting point closer to the optimal parameters and stabilizes the embedding space. There are declines when using the embedding transfer on \textsc{Entity}. This is because \textsc{Entity} contains massive new entities and needs more plasticity rather than stability. But on \textsc{Relation} and \textsc{Hybrid}, the results of ``w/o embedding transfer''  are lower than the full model, showing that embedding transfer can reduce the interference caused by the KG schema changes.
On \textsc{Fact}, the results of ``w/o embedding transfer'' are similar to the full model. 
This shows that, even without embedding transfer, the model can still capture the knowledge.
  
\section{Conclusion and Future Work}
This paper studies lifelong embedding learning for growing KGs. 
For better knowledge transfer and retention, we propose a lifelong KG embedding model consisting of masked KG autoencoder, embedding transfer, and embedding regularization. 
Experiments on new datasets show better link prediction accuracy, knowledge transfer capability, and learning efficiency of our model. 
In future, we plan to study lifelong embedding learning in long-tail and low-resource settings.

\section*{Acknowledgments} 
This work was supported by National Natural Science Foundation of China (No. 62272219).
 
%\clearpage
\bibliography{ref.bib}

\begin{thebibliography}{32}
\providecommand{\natexlab}[1]{#1}

\bibitem[{Bordes et~al.(2013)Bordes, Usunier, Garc{\'{i}}a{-}Dur{\'{a}}n,
  Weston, and Yakhnenko}]{TransE}
Bordes, A.; Usunier, N.; Garc{\'{i}}a{-}Dur{\'{a}}n, A.; Weston, J.; and
  Yakhnenko, O. 2013.
\newblock Translating Embeddings for Modeling Multi-relational Data.
\newblock In \emph{{NeurIPS}}, 2787--2795.

\bibitem[{Chen et~al.(2021)Chen, He, Wu, and Wang}]{TACT}
Chen, J.; He, H.; Wu, F.; and Wang, J. 2021.
\newblock Topology-Aware Correlations Between Relations for Inductive Link
  Prediction in Knowledge Graphs.
\newblock In \emph{{AAAI}}, 6271--6278.

\bibitem[{Daruna et~al.(2021)Daruna, Gupta, Sridharan, and Chernova}]{CKGE}
Daruna, A.~A.; Gupta, M.; Sridharan, M.; and Chernova, S. 2021.
\newblock Continual Learning of Knowledge Graph Embeddings.
\newblock \emph{{IEEE} Robotics Autom. Lett.}, 6(2): 1128--1135.

\bibitem[{Dettmers et~al.(2018)Dettmers, Minervini, Stenetorp, and
  Riedel}]{ConvE}
Dettmers, T.; Minervini, P.; Stenetorp, P.; and Riedel, S. 2018.
\newblock Convolutional {2D} Knowledge Graph Embeddings.
\newblock In \emph{{AAAI}}, 1811--1818.

\bibitem[{Galkin et~al.(2022)Galkin, Denis, Wu, and Hamilton}]{NodePiece}
Galkin, M.; Denis, E.~G.; Wu, J.; and Hamilton, W.~L. 2022.
\newblock NodePiece: Compositional and Parameter-Efficient Representations of
  Large Knowledge Graphs.
\newblock In \emph{{ICLR}}.

\bibitem[{Guo, Sun, and Hu(2019)}]{RSN}
Guo, L.; Sun, Z.; and Hu, W. 2019.
\newblock Learning to Exploit Long-term Relational Dependencies in Knowledge
  Graphs.
\newblock In \emph{{ICML}}, volume~97, 2505--2514.

\bibitem[{Hamaguchi et~al.(2017)Hamaguchi, Oiwa, Shimbo, and Matsumoto}]{MEAN}
Hamaguchi, T.; Oiwa, H.; Shimbo, M.; and Matsumoto, Y. 2017.
\newblock Knowledge Transfer for Out-of-Knowledge-Base Entities: A Graph Neural
  Network Approach.
\newblock In \emph{{IJCAI}}, 1802--1808.

\bibitem[{Hou et~al.(2022)Hou, Liu, Cen, Dong, Yang, Wang, and Tang}]{GraphMAE}
Hou, Z.; Liu, X.; Cen, Y.; Dong, Y.; Yang, H.; Wang, C.; and Tang, J. 2022.
\newblock {GraphMAE}: Self-Supervised Masked Graph Autoencoders.
\newblock \emph{CoRR}, abs/2205.10803: 1--11.

\bibitem[{Ji et~al.(2022)Ji, Pan, Cambria, Marttinen, and Yu}]{KG_survey}
Ji, S.; Pan, S.; Cambria, E.; Marttinen, P.; and Yu, P.~S. 2022.
\newblock A Survey on Knowledge Graphs: Representation, Acquisition, and
  Applications.
\newblock \emph{IEEE Trans. Neural Netw. Learn. Syst.}, 33(2): 494--514.

\bibitem[{Kipf and Welling(2017)}]{Kipf_GCN}
Kipf, T.~N.; and Welling, M. 2017.
\newblock Semi-Supervised Classification with Graph Convolutional Networks.
\newblock In \emph{ICLR}.

\bibitem[{Kirkpatrick et~al.(2017)Kirkpatrick, Pascanu, Rabinowitz, Veness,
  Desjardins, Rusu, Milan, Quan, Ramalho, Grabska{-}Barwinska, Hassabis,
  Clopath, Kumaran, and Hadsell}]{EWC}
Kirkpatrick, J.; Pascanu, R.; Rabinowitz, N.~C.; Veness, J.; Desjardins, G.;
  Rusu, A.~A.; Milan, K.; Quan, J.; Ramalho, T.; Grabska{-}Barwinska, A.;
  Hassabis, D.; Clopath, C.; Kumaran, D.; and Hadsell, R. 2017.
\newblock Overcoming Catastrophic Forgetting in Neural Networks.
\newblock \emph{PNAS}, 114(13): 3521--3526.

\bibitem[{Kou et~al.(2020)Kou, Lin, Liu, Li, Zhou, and Zhang}]{DiCGRL}
Kou, X.; Lin, Y.; Liu, S.; Li, P.; Zhou, J.; and Zhang, Y. 2020.
\newblock Disentangle-based Continual Graph Representation Learning.
\newblock In \emph{{EMNLP}}, 2961--2972.

\bibitem[{Liu et~al.(2021)Liu, Grau, Horrocks, and Kostylev}]{INDIGO}
Liu, S.; Grau, B.~C.; Horrocks, I.; and Kostylev, E.~V. 2021.
\newblock {INDIGO}: {GNN}-Based Inductive Knowledge Graph Completion Using
  Pair-Wise Encoding.
\newblock In \emph{{NeurIPS}}, 2034--2045.

\bibitem[{Lomonaco and Maltoni(2017)}]{CWR}
Lomonaco, V.; and Maltoni, D. 2017.
\newblock {CORe50}: A New Dataset and Benchmark for Continuous Object
  Recognition.
\newblock In \emph{CoRL}, volume~78, 17--26.

\bibitem[{Lopez{-}Paz and Ranzato(2017)}]{GEM}
Lopez{-}Paz, D.; and Ranzato, M. 2017.
\newblock Gradient Episodic Memory for Continual Learning.
\newblock In \emph{{NeurIPS}}, 6467--6476.

\bibitem[{Rusu et~al.(2016)Rusu, Rabinowitz, Desjardins, Soyer, Kirkpatrick,
  Kavukcuoglu, Pascanu, and Hadsell}]{PNN}
Rusu, A.~A.; Rabinowitz, N.~C.; Desjardins, G.; Soyer, H.; Kirkpatrick, J.;
  Kavukcuoglu, K.; Pascanu, R.; and Hadsell, R. 2016.
\newblock Progressive Neural Networks.
\newblock \emph{CoRR}, abs/1606.04671: 1--14.

\bibitem[{Scarselli et~al.(2009)Scarselli, Gori, Tsoi, Hagenbuchner, and
  Monfardini}]{GCN}
Scarselli, F.; Gori, M.; Tsoi, A.~C.; Hagenbuchner, M.; and Monfardini, G.
  2009.
\newblock The Graph Neural Network Model.
\newblock \emph{IEEE Trans. Neural Netw.}, 20(1): 61--80.

\bibitem[{Schlichtkrull et~al.(2018)Schlichtkrull, Kipf, Bloem, van~den Berg,
  Titov, and Welling}]{R-GCN}
Schlichtkrull, M.; Kipf, T.~N.; Bloem, P.; van~den Berg, R.; Titov, I.; and
  Welling, M. 2018.
\newblock Modeling Relational Data with Graph Convolutional Networks.
\newblock In \emph{{ESWC}}, 593--607.

\bibitem[{Tay, Luu, and Hui(2017)}]{puTransE}
Tay, Y.; Luu, A.~T.; and Hui, S.~C. 2017.
\newblock Non-Parametric Estimation of Multiple Embeddings for Link Prediction
  on Dynamic Knowledge Graphs.
\newblock In \emph{AAAI}, 1243--1249.

\bibitem[{Teru, Denis, and Hamilton(2020)}]{GraIL}
Teru, K.; Denis, E.; and Hamilton, W. 2020.
\newblock Inductive Relation Prediction by Subgraph Reasoning.
\newblock In \emph{ICML}, 9448--9457.

\bibitem[{Toutanova and Chen(2015)}]{FB237}
Toutanova, K.; and Chen, D. 2015.
\newblock Observed Versus Latent Features for Knowledge Base and Text
  Inference.
\newblock In \emph{CVSC}, 57--66.

\bibitem[{Vashishth et~al.(2020{\natexlab{a}})Vashishth, Sanyal, Nitin,
  Agrawal, and Talukdar}]{InteractE}
Vashishth, S.; Sanyal, S.; Nitin, V.; Agrawal, N.; and Talukdar, P.~P.
  2020{\natexlab{a}}.
\newblock InteractE: Improving Convolution-based Knowledge Graph Embeddings by
  Increasing Feature Interactions.
\newblock In \emph{AAAI}, 3009--3016.

\bibitem[{Vashishth et~al.(2020{\natexlab{b}})Vashishth, Sanyal, Nitin, and
  Talukdar}]{CompGCN}
Vashishth, S.; Sanyal, S.; Nitin, V.; and Talukdar, P.~P. 2020{\natexlab{b}}.
\newblock Composition-based Multi-Relational Graph Convolutional Networks.
\newblock In \emph{ICLR}.

\bibitem[{Vaswani et~al.(2017)Vaswani, Shazeer, Parmar, Uszkoreit, Jones,
  Gomez, Kaiser, and Polosukhin}]{Transformer}
Vaswani, A.; Shazeer, N.; Parmar, N.; Uszkoreit, J.; Jones, L.; Gomez, A.~N.;
  Kaiser, L.; and Polosukhin, I. 2017.
\newblock Attention is All you Need.
\newblock In \emph{NeurIPS}, 5998--6008.

\bibitem[{Vrandecic and Kr{\"{o}}tzsch(2014)}]{Wikidata}
Vrandecic, D.; and Kr{\"{o}}tzsch, M. 2014.
\newblock Wikidata: A Free Collaborative Knowledgebase.
\newblock \emph{Commun. ACM}, 57(10): 78--85.

\bibitem[{Wang et~al.(2019{\natexlab{a}})Wang, Xiong, Yu, Guo, Chang, and
  Wang}]{EMR}
Wang, H.; Xiong, W.; Yu, M.; Guo, X.; Chang, S.; and Wang, W.~Y.
  2019{\natexlab{a}}.
\newblock Sentence Embedding Alignment for Lifelong Relation Extraction.
\newblock In \emph{NAACL}, 796--806.

\bibitem[{Wang et~al.(2019{\natexlab{b}})Wang, Han, Li, and Pan}]{LAN}
Wang, P.; Han, J.; Li, C.; and Pan, R. 2019{\natexlab{b}}.
\newblock Logic Attention Based Neighborhood Aggregation for Inductive
  Knowledge Graph Embedding.
\newblock In \emph{AAAI}, 7152--7159.

\bibitem[{Wang et~al.(2017)Wang, Mao, Wang, and Guo}]{KGE_survey}
Wang, Q.; Mao, Z.; Wang, B.; and Guo, L. 2017.
\newblock Knowledge Graph Embedding: {A} Survey of Approaches and Applications.
\newblock \emph{IEEE Trans. Knowl. Data Eng.}, 29(12): 2724--2743.

\bibitem[{Wang et~al.(2014)Wang, Zhang, Feng, and Chen}]{TransH}
Wang, Z.; Zhang, J.; Feng, J.; and Chen, Z. 2014.
\newblock Knowledge Graph Embedding by Translating on Hyperplanes.
\newblock In \emph{AAAI}, 1112--1119.

\bibitem[{Wu et~al.(2022)Wu, Khan, Yong, Qi, and Wang}]{DKGE}
Wu, T.; Khan, A.; Yong, M.; Qi, G.; and Wang, M. 2022.
\newblock Efficiently Embedding Dynamic Knowledge Graphs.
\newblock \emph{Knowl. Based Syst.}, 250: 109124.

\bibitem[{Zenke, Poole, and Ganguli(2017)}]{SI}
Zenke, F.; Poole, B.; and Ganguli, S. 2017.
\newblock Continual Learning through Synaptic Intelligence.
\newblock In \emph{ICML}, 3987--3995.

\bibitem[{Zhu et~al.(2021)Zhu, Zhang, Xhonneux, and Tang}]{NBFNet}
Zhu, Z.; Zhang, Z.; Xhonneux, L. A.~C.; and Tang, J. 2021.
\newblock Neural Bellman-Ford Networks: {A} General Graph Neural Network
  Framework for Link Prediction.
\newblock In \emph{NeurIPS}, 29476--29490.

\end{thebibliography}

\end{document}